\documentclass{article}
\usepackage{ijcai18}

\usepackage{times}
\usepackage{xcolor}
\usepackage{url}
\usepackage{soul}
\usepackage[hidelinks]{hyperref}
\usepackage[utf8]{inputenc}
\usepackage[small]{caption}

\usepackage{CJKutf8}

\usepackage{booktabs} 

\usepackage{graphicx}
\usepackage{bbm}
\usepackage{subfigure}
\usepackage{courier}
\usepackage{color, colortbl}
\usepackage{verbatim}
\usepackage[linesnumbered,ruled,vlined]{algorithm2e}
\usepackage{amssymb,amsmath}
\usepackage{bm}

\title{A Reinforced Topic-Aware Convolutional Sequence-to-Sequence\\ Model for Abstractive Text Summarization}

\author{
Li Wang$^1$, 
Junlin Yao$^2$, 
Yunzhe Tao$^3$, 
Li Zhong$^1$,
Wei Liu$^4$,
Qiang Du$^3$
\\ 
$^1$ Tencent Data Center of SNG\\
$^2$ ETH Zürich\\
$^3$ Columbia University\\
$^4$ Tencent AI Lab  \\
lilianwang@tencent.com,
jyao@student.ethz.ch,
y.tao@columbia.edu, \\
reggiezhong@tencent.com,
wl2223@columbia.edu,
qd2125@columbia.edu
}
\begin{document}

\maketitle

\begin{abstract}
In this paper, we propose a deep learning approach to tackle the automatic summarization tasks by incorporating topic information into the convolutional sequence-to-sequence (ConvS2S) model and using self-critical sequence training (SCST) for optimization. Through jointly attending to topics and word-level alignment, our approach can improve coherence, diversity, and informativeness of generated summaries via a biased probability generation mechanism. On the other hand, reinforcement training, like SCST, directly optimizes the proposed model with respect to the non-differentiable metric ROUGE, which also avoids the exposure bias during inference. We carry out the experimental evaluation with state-of-the-art methods over the Gigaword, DUC-2004, and LCSTS datasets. The empirical results demonstrate the superiority of our proposed method in the abstractive summarization.
\end{abstract}

%
%



\section{Introduction}
Automatic text summarization has played an important role in a variety of natural language processing (NLP) applications, such as news headlines generation \cite{kraaij2002headline} and feeds stream digests \cite{barzilay2005sentence}. 
It is of interest to generate informative and representative natural language summaries which are capable of retaining the main ideas of source articles. The key challenges in automatic text summarization are correctly evaluating and selecting important information, efficiently filtering redundant contents, and properly aggregating related segments and making human-readable summaries. 
Compared to other NLP tasks, the automatic summarization has its own difficulties. 
For example, unlike machine translation tasks where input and output sequences often share similar lengths, summarization tasks are more likely to have input and output sequences greatly imbalanced. Besides, machine translation tasks usually have some direct word-level alignment between input and output sequences, which is less obvious in summarization. 
%
%


There are two genres of automatic summarization techniques, namely extraction and abstraction.
The goal of extractive summarization \cite{neto2002automatic} is to produce a summary by selecting important pieces of the source document and concatenating them verbatim, 
while abstractive summarization \cite{chopra2016abstractive} generates summaries based on the core ideas of the document, therefore the summaries could be paraphrased in more general terms. 
Other than extraction, abstractive methods should be able to 
properly rewrite the core ideas of the source document and assure that the generated summaries are grammatically correct and human readable, which is close to the way how humans do summarization and thus is of interest to us in this paper. 

Recently, deep neural network models have been widely used for NLP tasks such as machine translation \cite{bahdanau2014neural}, 
and text summarization \cite{nallapati2016abstractive}. In particular, the attention based sequence-to-sequence framework \cite{bahdanau2014neural} with recurrent neural networks (RNNs) \cite{sutskever2014sequence} 
prevails in the NLP tasks. However, RNN-based models are more prone to gradient vanishing due to their chain structure of non-linearities compared to the hierarchical structure of CNN-based models \cite{dauphin2016language}. In addition, the temporal dependence among the hidden states of RNNs prevents parallelization over the elements of a sequence, which makes the training inefficient.
%
%

In this paper, we propose a new approach based on the convolutional sequence-to-sequence (ConvS2S) framework \cite{gehring2017convolutional} jointly with a topic-aware attention mechanism. To the best of our knowledge, this is the first work for automatic abstractive summarization that incorporates the topic information, which can provide themed and contextual alignment information into deep learning architectures. 
%
%
In addition, we also optimize our proposed model 
by employing the reinforcement training \cite{DBLP:journals/corr/PaulusXS17}. The main contributions of this paper include:
\begin{itemize}
\item We propose a joint attention and biased probability generation mechanism to incorporate the topic information into an automatic summarization model, which introduces contextual information to help the model generate more coherent summaries with increased diversity and informativeness. 
\item We employ the self-critical sequence training technique in ConvS2S to directly optimize the model with respect to the non-differentiable summarization metric ROUGE, which 
also remedies the exposure bias issue.
\item Extensive experimental results on three benchmark datasets demonstrate that by fully exploiting the power of the ConvS2S architecture enhanced by topic embedding and SCST, our proposed model yields high accuracy for abstractive summarization, advancing the state-of-the-art methods. 
\end{itemize}

\section{Related Work}
Automatic text summarization has been widely investigated. Many approaches have been proposed to address this challenging task. 
Various methods \cite{neto2002automatic} focus on the extractive summarization, which select important contents of text and combine them verbatim to produce a summary. On the other hand, abstractive summarization models are able to produce a grammatical summary with a novel expression, most of which \cite{rush2015neural,chopra2016abstractive,nallapati2016sequence} are built upon the neural attention-based sequence-to-sequence framework \cite{sutskever2014sequence}.

The predominant models are based on the RNNs \cite{nallapati2016abstractive,shen2016neural,DBLP:journals/corr/PaulusXS17}, where the encoder and decoder are constructed using either Long Short-Term Memory (LSTM) \cite{hochreiter1997long} or Gated Recurrent Unit (GRU) \cite{cho2014learning}. However, very few methods have explored the performance of convolutional structure in summarization tasks. Compared to RNNs, convolutional neural networks (CNNs) enjoy several advantages, including the efficient training by leveraging parallel computing, and mitigating the gradient vanishing problem due to fewer non-linearities \cite{dauphin2016language}. 
Notably, the recently proposed gated convolutional network \cite{dauphin2016language,gehring2017convolutional} outperforms state-of-the-art RNN-based models in the language modeling and machine translation tasks.
%
%
%

While the ConvS2S model is also evaluated on the abstractive summarization \cite{gehring2017convolutional}, there are several limitations. First, the model is trained by minimizing a maximum-likelihood loss which is sometimes inconsistent with the quality of a summary and the metric that is evaluated from the whole sentences, such as ROUGE~\cite{lin2004rouge}
%
%
In addition, the exposure bias \cite{ranzato2015sequence} occurs due to only exposing the model to the training data distribution instead of its own predictions. 
%
%
%
More importantly, the ConvS2S model utilizes only word-level alignment which may be insufficient for summarization and prone to incoherent generalized summaries. Therefore, the higher level alignment could be a potential assist. For example, the topic information has been introduced to a RNN-based sequence-to-sequence model \cite{xing2017topic} for chatbots to generate more informative responses.


\section{Reinforced Topic-Aware Convolutional Sequence-to-Sequence Model}
\begin{figure}[t]
\centering
\includegraphics[width=1\columnwidth]{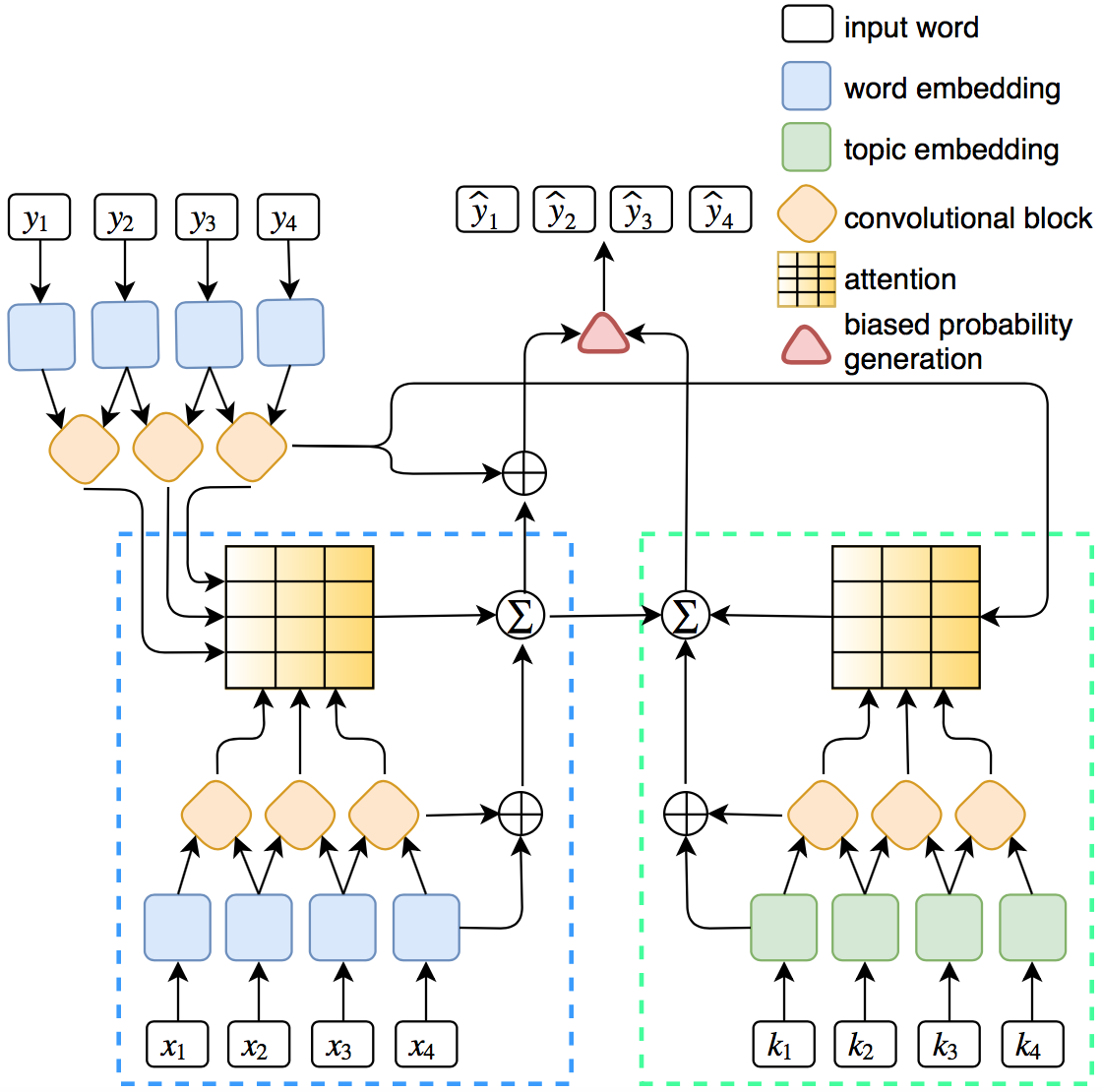}
\caption{A graphical illustration of the topic-aware convolutional architecture. Word and topic embeddings of the source sequence are encoded by the associated convolutional blocks (bottom left and bottom right). Then we jointly attend to words and topics by computing dot products of decoder representations (top left) and word/topic encoder representations. Finally, we produce the target sequence through a biased probability generation mechanism.}
\label{fig:architecture}
\end{figure}

In this section, we propose the Reinforced Topic-Aware Convolutional Sequence-to-Sequence model, which consists of a convolutional architecture with both input words and topics, a joint multi-step attention mechanism, a biased generation structure, and a reinforcement learning procedure. The graphical illustration of the topic-aware convolutional architecture can be found in Figure \ref{fig:architecture}. 

\subsection{ConvS2S Architecture}
We exploit the ConvS2S architecture \cite{gehring2017convolutional} as the basic infrastructure of our model. In this paper, two convolutional blocks are employed, associated with the word-level and topic-level embeddings, respectively. We introduce the former in this section and the latter in next, along with the new joint attention and the biased generation mechanism.

\subsubsection{Position Embeddings}
Let $\bm{x}=(x_1,\ldots,x_m)$ denote the input sentence. We first embed the input elements (words) in a distributional space as $\bm{w}=(w_1,\ldots,w_m)$, where $w_i\in\mathbb{R}^d$ are rows of a randomly initialized matrix $\mathcal{D}_\text{word}\in\mathbb{R}^{V\times d}$ with $V$ being the size of vocabulary. We also add a positional embedding, $\bm{p}=(p_1,\ldots,p_m)$ with $p_i\in\mathbb{R}^d$, to retain the order information. Thus, the final embedding for the input is $\bm{e}=(w_1+p_1,\ldots,w_m+p_m)$. Similarly, let $\bm{q}=(q_1,\ldots,q_n)$ denote the embedding for output elements that were already generated by the decoder and being fed back to the next step.
%
%
%

\subsubsection{Convolutional Layer}
\label{sec:convlayer}
Both encoder and decoder networks are built by stacking several convolutional layers.
Suppose that the kernel has width of $k$ and the input embedding dimension is $d$. The convolution takes a concatenation of $k$ input elements $X\in\mathbb{R}^{kd}$ and maps it to an output element $Y\in\mathbb{R}^{2d}$, namely, 
\begin{equation}
Y = f_{\text{conv}}(X) \doteq W_Y X+b_Y\,,
\end{equation}
where the kernel matrix $W_Y \in \mathbb{R}^{2d \times kd}$ and the bias term $b_Y \in \mathbb{R}^{2d}$ are the parameters to be learned. 

Rewrite the output as $Y = [A; B]$, where $A, B\in \mathbb{R}^{d}$. Then the gated linear unit (GLU) \cite{dauphin2016language} is given by 
\begin{equation}
g([A; B]) = A \otimes \sigma(B)\,,
\end{equation}
where $\sigma$ is the sigmoid function, $\otimes$ is the point-wise multiplication, and the output of GLU is in $\mathbb{R}^d$.

We denote the outputs of the $l$-th layer as $\bm{h}^l=(h_1^l,\ldots,h_n^l)$ for the decoder, and $\bm{z}^l=(z_1^l,\ldots,z_m^l)$ for the encoder. 
Take the decoder for illustration. The convolution unit $i$ on the $l$-th layer is computed by residual connections as
\begin{equation}
h_i^l = g\circ f_{\text{conv}}\left(
\begin{bmatrix}
h_{i-k/2}^{l-1}; \cdots; h_{i+k/2}^{l-1}
\end{bmatrix}
\right)+h_i^{l-1}\,,
\end{equation}
where $h^l_i \in \mathbb{R}^d$ and $\circ$ is the function composition operator.

\subsubsection*{Multi-step Attention}
The attention mechanism is introduced to make the model access historical information. 
To compute the attention, we first embed the current decoder state $h_i^l$ as
\begin{equation}\label{eq:em}
d_i^l=W_d^l h_i^l+b_d^l+q_i\,,
\end{equation}
where $q_i\in\mathbb{R}^d$ is the embedding of the previous decoded element. Weight matrix $W_d^l\in\mathbb{R}^{d\times d}$ and bias $b_d^l\in\mathbb{R}^d$ are the parameters to be learned.  

The attention weights $\alpha{_{ij}^l}$ of state $i$ and source input element $j$ is computed as a dot product between $d{_i^l}$ and the output $z{_j^{u_o}}$ of the last encoder block $u_o$, namely,
\begin{equation}
\alpha{_{ij}^l}=\frac{\exp(d{_i^l} \cdot z{_j^{u_o}})}{\sum\limits_{t = 1}^m{\exp(d{_i^l} \cdot z{_t^{u_o}})} }\,.
\label{eq:aold}
\end{equation}

The conditional input ${c_i^l}\in\mathbb{R}^d$ for the current decoder layer is computed as
\begin{equation}
c{_i^l}=\sum\limits_{j=1}^m{\alpha{_{ij}^l}(z{_j^{u_o}}+e_j)}\,,
\label{eq:c}
\end{equation}
where $e_j$ is the input element embedding that can provide point information about a specific input element. Once $c{_i^l}$ has been computed, it is added to the output of the corresponding decoder layer $h{_i^l}$ and serves as a part of the input to $h_i^{l+1}$. 

\subsection{Topic-Aware Attention Mechanism}
A topic model is a type of statistical model for discovering the abstract ideas or hidden semantic structures that occur in a collection of source articles. In this paper, we employ the topic model to acquire latent knowledge of documents and incorporate a topic-aware mechanism into the multi-step attention-based ConvS2S model, which is expected to bring prior knowledge for text summarization. 
Now we present the novel approach on how to incorporate the topic model into the basic ConvS2S framework via the joint attention mechanism and biased probability generation process.

\subsubsection*{Topic Embeddings}
The topic embeddings are obtained by classical topic models such as Latent Dirichlet Allocation (LDA) \cite{blei2003latent}. During pre-training, we use LDA to assign topics to the input texts. The top $N$ non-universal words with the highest probabilities of each topic are chosen into the topic vocabulary $\bm{K}$. More details will be given in Section \ref{sec:setup}. While the vocabulary of texts is denoted as $\bm{V}$, we assume that $\bm{K} \subset \bm{V}$. Given an input sentence $\bm{x}=(x_1,\ldots,x_m)$, if a word $x_i\notin\bm{K}$, we embed it as before to attain $w_i$. However, if a word $x_i\in\bm{K}$, we can embed this topic word as $t_i\in\mathbb{R}^d$, which is a row in the topic embedding matrix $\mathcal{D}_\text{topic} \in\mathbb{R}^{K\times d}$, where $K$ is the size of topic vocabulary. The embedding matrix $\mathcal{D}_\text{topic}$ is normalized from the corresponding pre-trained topic distribution matrix, whose row is proportional to the number of times that each word is assigned to each topic. In this case, the positional embedding vectors 
are also added to the encoder and decoder elements, respectively, to obtain the final topic embeddings $\bm{r}=(r_1,\ldots,r_m)$ and $\bm{s}=(s_1,\ldots,s_n)$.

\subsubsection*{Joint Attention}
Again we take the decoder for illustration. Following the convolutional layer introduced before, we can obtain the convolution unit $i$ on the $l$-th layer in the decoder of topic level as $\tilde{h}_i^l\in\mathbb{R}^d$. Similar to \eqref{eq:em}, we have
\begin{equation}
\tilde{d}_i^l = \tilde{W}_d^l \tilde{h}_i^l + \tilde{b}_d^l + {s}_i\,.
\end{equation}
We then incorporate the topic information into the model through a joint attention mechanism. During decoding, the joint attention weight $\beta_{ij}^{l}$ is given by
\begin{equation}
\beta{_{ij}^{l}}=\frac{\exp(\tilde{d}{_i^{l}} \cdot z{_j^{u_o}} + \tilde{d}{_i^{l}} \cdot z{_j^{u_t}})}{\sum\limits_{t = 1}^m{\exp(\tilde{d}{_i^{l}} \cdot z{_t^{u_o}} + \tilde{d}{_i^{l}} \cdot z{_t^{u_t}} )} }\,,
\label{eq:a}
\end{equation}
where 
$z{_j^{u_t}}$ is the output of the last topic-level encoder block $u_t$. 
Then the conditional input $\tilde{c}_i^l\in\mathbb{R}^d$ is computed as
\begin{equation}
\tilde{c}{_i^l}=\sum\limits_{j=1}^m{\beta{_{ij}^l}(z{_j^{u_t}}+r_j)}\,.
\end{equation}
In the joint attention mechanism, both $\tilde{c}_i^l$ and $c_i^l$ are added to the output of the corresponding decoder layer $\tilde{h}_i^l$ and are a part of the input to $\tilde{h}_i^{l+1}$.

\subsubsection*{Biased Probability Generation}
Finally, we compute a distribution over all possible next target elements $y_{i+1}\in\mathbb{R}^T$, namely 
\begin{equation}\label{eq:prob}
p_\theta(y_{i+1}):= p(y_{i+1}|y_1,\ldots,y_i,\bm{x})\in\mathbb{R}^T\,,
\end{equation}
by transforming the top word-level decoder outputs $h^{L_o}$ and topic-level decoder outputs $\tilde{h}^{L_t}$ via a linear layer $\Psi(\cdot)$, which is computed by
\begin{equation}
\Psi(h) = W_o h + b_o\,,
\end{equation}
where $W_o\in\mathbb{R}^{T\times d}$ and $b_o\in\mathbb{R}^T$ are the parameters to be learned. Then the biased generation distribution is given as
\begin{equation}
p_\theta(y_{i+1})= \frac{1}{Z} \left[ \exp\left(\Psi(h{_i^{L_o}})\right) + \exp\left(\Psi(\tilde{h}{_i^{L_t}})\right)\otimes I_{\{w\in\bm{K}\} }\right] \,, 
\label{eq:py}
\end{equation}
where $Z$ is the normalizer, $h{_i^{L_o}}$ and $\tilde{h}{_i^{L_t}}$ denote the $i$-th top decoder outputs of word and topic, respectively, and $I$ is the one-hot indicator vector of each candidate word $w$ in $y_{i+1}$. When the candidate word $w$ is a topic word, we bias the generation distribution by the topic information. Otherwise, we ignore the topic part. To some extent, the complexity of the search space is reduced by introducing the topic bias since important words are more likely to  be generated directly.

\subsection{Reinforcement Learning}
The teacher forcing algorithm \cite{williams1089a} 
aims to minimize the maximum-likelihood loss at each decoding step, namely,
\begin{equation}
\label{eq:ml}
L_{\text{ml}} = - \sum_{i = 1}^T \log p_\theta (y^{\ast}_i | y^{\ast}_1, y^{\ast}_2,\ldots, y^{\ast}_{i -
   1}, \bm{x})\,,
\end{equation}
where $\bm{x}$ refers to an input sequence and $y^{\ast}=(y^{\ast}_1,$$y^{\ast}_2,$$\ldots,$$y^{\ast}_T)$ is the corresponding ground-truth output sequence. 

Minimizing the objective in Eq.~\eqref{eq:ml} often produces sub-optimal results with respect to the evaluation metrics, such as ROUGE which measures the sentence-level accuracy of the generated summaries. 
The sub-optimality is related to the problem called {\em exposure bias} \cite{ranzato2015sequence}, which is caused by only exposing a model to the distribution of training data instead of its own distribution. During the training process, models are fed by ground-truth output sequences to predict the next word, whereas during inference they generate the next word given the predicted words as inputs. Therefore, in the test process, the error of each step accumulates and leads to the deterioration of performance. 

The second reason for sub-optimality comes from the flexibility of summaries. The maximum-likelihood objective rewards models that can predict exactly the same summaries as references while penalizing those that produce different texts even though they are semantically similar.
Providing multiple reference summaries is helpful yet insufficient since there are alternatives to rephrase a given summary.
Therefore, minimizing the objective in Eq. \eqref{eq:ml} neglects the intrinsic property of summarization. ROUGE, on the other hand, provides more flexible evaluation, encouraging models to focus more on semantic meanings than on word-level correspondences.

In order to address such issues, we utilize self-critical sequence training (SCST) \cite{rennie2016self}, a policy gradient algorithm for reinforcement learning, to directly maximize the non-differentiable ROUGE metric. During reinforcement learning, we generate two output sequences given the input sequence $\bm{x}$. The first sequence $\hat{y}$ is obtained by greedily selecting words that maximize the output probability distribution, and the other output sequence $y^s$ is generated by sampling from the distribution. After obtaining ROUGE scores of both sequences as our rewards, i.e., $r(y^s)$ and $r(\hat{y})$, we minimize the reinforcement loss
\begin{equation}
L_\text{rl} = -(r (y^s) -
   r (\hat{y})) \log p_{\theta} (y^s),
\end{equation}
and update model parameters by gradient descent techniques.

\begin{table}[!ht]
\centering
	\resizebox{\columnwidth}{!}{
	{
	\begin{tabular}{c|l}
    \toprule\hline
		 \textbf{No.} & \multicolumn{1}{c}{\textbf{Topic Words}}\\
         \hline
         1 & prime, minister, talks, leader, elections, visit \\
		 2 & bird, flu, officials, opens, poultry, die \\
		 3 & trade, free, EU, army, urges, ban\\
         4 & Bush, world, talks, foreign, investment, markets\\
         5 & world, Malaysia, Thailand, meet, Vietnam, U.S.\\
         \hline
		\bottomrule
	\end{tabular}
	}}
    \caption{Examples of topic words for the Gigaword corpus.}
    \label{tab:topic_words_giga}    
\end{table}

With SCST, we can directly optimize the discrete evaluation metric. In addition, the ``self-critical'' test-time estimate of the reward $r(\hat{y})$ provides a simple yet effective baseline and improves training/test time consistency. Since during learning we set the baseline of the REINFORCE algorithm as the reward obtained by the current model in the test-time inference, the SCST exposes the model to its own distribution and encourages it to produce the sequence output $\hat{y}$ with a high ROUGE score, avoiding the exposure bias issue and thus improving the test performance.

\section{Experimental Setup}\label{sec:setup}
\subsection{Datasets}
In this paper, we consider three datasets to evaluate the performance of different methods in the abstractive text summarization task. First, we consider the annotated Gigaword corpus \cite{graff2003english} preprocessed identically to \cite{rush2015neural}, which leads to around 3.8M training samples, 190K validation samples and 1951 test samples for evaluation. The input summary pairs consist of the headline and the first sentence of the source articles. 
We also evaluate various models on the DUC-2004 test set\footnote{\url{http://duc.nist.gov/data.html}}~\cite{over2007duc}.
The dataset is a standard summarization evaluation set, which consists of 500 news articles. 
Unlike the Gigaword corpus, each article in DUC-2004 is paired with four human-generated reference summaries, which makes the evaluation more objective. 
The last dataset for evaluation is a large corpus of Chinese short text summarization (LCSTS) dataset \cite{hu2015lcsts} collected and constructed from the Chinese microblogging website Sina Weibo. Following the setting in the original paper, we use the first part of LCSTS dataset for training, which contains 2.4M text-summary pairs, and choose 725 pairs from the last part with high annotation scores as our test set.


\subsection{Topic Information}
The classical LDA with Gibbs Sampling technique is used to pre-train the corpus for topic embedding initialization and provide candidates for the biased probability generation process.
%
%
The topic embedding values are normalized to a distribution with mean zero and variance of 0.1 for adaption to the neural network structure. In this paper, we pick top $N=200$ words with the highest probabilities in each topic to obtain the topic word set. Note that the universal words are filtered out during pre-training.
Randomly selected examples of topic words of the Gigaword corpus are presented in Table \ref{tab:topic_words_giga}.

\begin{table}[htbp]
\centering
	\resizebox{\columnwidth}{!}{
	{
	\begin{tabular}{l|c|c|c}
		\toprule
        \hline
		 & \textbf{RG-1 (F)} & \textbf{RG-2 (F)} & \textbf{RG-L (F)}\\
		\hline
        ABS \cite{rush2015neural} & 29.55 & 11.32 & 26.42 \\
		ABS+ \cite{rush2015neural} & 29.76 & 11.88 & 26.96 \\
		RAS-Elman \cite{chopra2016abstractive} & 33.78 & 15.97 & 31.15\\
		words-lvt5k-1sent \cite{nallapati2016abstractive} & 35.30 &16.64 & 32.62\\ 
        RNN+MLE \cite{shen2016neural}  &32.67 & 15.23 & 30.56 \\
        RNN+MRT \cite{shen2016neural}  & 36.54 & 16.59 & 33.44 \\
        SEASS(beam) \cite{zhou2017selective}  &36.15 & 17.54 & 33.63 \\
        ConvS2S \cite{gehring2017convolutional} &35.88 & 17.48 & 33.29 \\
        \hline
        Topic-ConvS2S & 36.38 & 17.96 & 34.05 \\
        Reinforced-ConvS2S & 36.30 & 17.64 & 33.90 \\
        Reinforced-Topic-ConvS2S & \textbf{36.92} & $\textbf{18.29}$ & $\textbf{34.58}$ \\ 
        \hline
		\bottomrule
	\end{tabular}
	}}
    \caption{Accuracy on the Gigaword corpus in terms of the full-length ROUGE-1 (RG-1), ROUGE-2 (RG-2), and ROUGE-L (RG-L). Best performance on each score is displayed in \textbf{boldface}.}
    \label{tab:res_giga1}
\end{table}

\begin{table}[htbp]
    \centering
	\resizebox{\columnwidth}{!}{
	{
	\begin{tabular}{l|c|c|c}
		\toprule
        \hline
		 & \textbf{RG-1 (F)} & \textbf{RG-2 (F)} & \textbf{RG-L (F)}\\
		\hline
        ABS (beam) \cite{rush2015neural} & 37.41 & 15.87 & 34.70 \\
		s2s+att (greedy) \cite{zhou2017selective} & 42.41 & 20.76 & 39.84 \\
		s2s+att (beam) \cite{zhou2017selective} & 43.76 & 22.28 & 41.14\\
		SEASS (greedy) \cite{zhou2017selective} & 45.27 &22.88 & 42.20\\ 
        SEASS (beam) \cite{zhou2017selective}  &46.86 & 24.58 & 43.53 \\
        \hline
        Topic-ConvS2S & 46.80 & 24.74 & 43.92 \\
        Reinforced-ConvS2S & 46.68 &24.22  & 43.76 \\
        Reinforced-Topic-ConvS2S & $\textbf{46.92}$ & $\textbf{24.83}$ & $\textbf{44.04}$ \\     
        \hline
		\bottomrule
	\end{tabular}
	}}
    \caption{Accuracy on the internal test set of Gigaword corpus in terms of the full-length RG-1, RG-2, and RG-L. Best performance on each score is displayed in \textbf{boldface}.}
    \label{tab:res_giga2}
\end{table}

\subsection{Model Parameters and Optimization}
We employ six convolutional layers for both the encoder and decoder. 
All embeddings, including the initialized embedding and the output produced by the decoder before the final linear layer, have a dimensionality of 256. We also adopt the same dimensionality for the size of linear layer mapping between hidden and embedding states. 
We use a learning rate of 0.25 and reduce it by a decay rate of 0.1 once the validation ROUGE score stops increasing after each epoch until the learning rate falls below $10^{-5}$. 
We first train the basic topic-aware convolutional model with respect to a standard maximum likelihood objective, and then switch to further minimize a mixed training objective \cite{DBLP:journals/corr/PaulusXS17}, incorporating the reinforcement learning objective $L_\text{rl}$ and the original maximum likelihood $L_\text{ml}$, which is given as
\begin{equation}
L_\text{mixed} = \lambda L_\text{rl} + (1 - \lambda)L_\text{ml},
\end{equation}
where the scaling factor $\lambda$ is set to be $0.99$ in our experiments. Moreover, we choose the ROUGE-L metric as the reinforcement reward function. Nesterov's accelerated gradient method \cite{sutskever2013importance} is used for training, with the mini-batch size of $32$ and the learning rate of $0.0001$.
All models are implemented in PyTorch \cite{paszke2017pytorch} and trained on a single Tesla M40 GPU.

\begin{table}[htbp]
	\centering
	\resizebox{\columnwidth}{!}{
	{
	\begin{tabular}{m{10cm}}
		\toprule
        \hline
		 \textbf{Examples of summaries} \\
         \hline   
         
         
         $\textbf{D}$: the sri lankan government on wednesday announced the closure of government schools with immediate effect as a military campaign against tamil separatists escalated in the north of the country.\\
         $\textbf{R}$: sri lanka closes schools as war escalates \\
         $\textbf{OR}$: sri lanka closes schools with immediate {\color{blue}effect} \\
         $\textbf{OT}$: sri lanka closes schools in {\color{red}wake} of {\color{blue}military} {\color{red}attacks}\\
         \hline
         
         $\textbf{D}$: a us citizen who spied for communist east germany was given a suspended jail sentence of \#\# months here friday.\\
         $\textbf{R}$: us citizen who spied for east germans given suspended sentence \\
         $\textbf{OR}$: us man gets suspended {\color{blue}jail} {\color{red}term} for {\color{blue}communist spying} \\
         $\textbf{OT}$: us man {\color{blue}jailed} for {\color{red}espionage}\\
         \hline
         
         $\textbf{D}$: malaysian prime minister mahathir mohamad indicated he would soon relinquish control of the ruling party to his deputy anwar ibrahim.\\
         $\textbf{R}$: mahathir wants leadership change to be smooth \\
         $\textbf{OR}$: malaysia's mahathir to relinquish {\color{blue}control of ruling party} \\
         $\textbf{OT}$: malaysia's mahathir to {\color{red}submit} {\color{blue}control of ruling party}\\
         \hline
         
         $\textbf{D}$: a french crocodile farm said it had stepped up efforts to breed one of the world's most endangered species, the indian UNK, with the hope of ultimately returning animals to their habitat in south asia.\\
         $\textbf{R}$: french farm offers hope for endangered asian crocs UNK picture \\
         $\textbf{OR}$: french crocodile farm {\color{blue}steps} up {\color{blue}efforts} to breed endangered {\color{blue}species} \\
         $\textbf{OT}$: french crocodile farm says {\color{blue}steps} up {\color{blue}efforts} to {\color{red}save} endangered {\color{blue}species}\\
         \hline
         
         
		\bottomrule
	\end{tabular}
	}}
    \caption{Examples of generated summaries on the Gigaword corpus. \textbf{D}: source document, \textbf{R}: reference summary, \textbf{OR}: output of the \textbf{R}einforced-ConvS2S model, \textbf{OT}: output of the Reinforced-\textbf{T}opic-ConvS2S model. The words marked in {\color{blue}blue} are topic words not in the reference summaries. The words marked in {\color{red}red} are topic words neither in the reference summaries nor in the source documents.}
    \label{tab:sum_giga}
\end{table}

\section{Results and Analysis}
We follow the existing work and adopt the ROUGE metric \cite{lin2004rouge} for evaluation.
\subsection{Gigaword Corpus}

We demonstrate the effectiveness of our proposed model via a step-by-step justification. First, the basic ConvS2S structure with topic-aware model or reinforcement learning is tested, respectively. Then we combine the two to show the performance of our Reinforced-Topic-ConvS2S model. We report the full-length F-1 scores of the ROUGE-1 (RG-1), ROUGE-2 (RG-2), and ROUGE-L (RG-L) metrics and compare the results with various neural abstractive summarization methods, which are presented in Table \ref{tab:res_giga1}. The ABS and ABS+ models are attention-based neural models for text summarization. The RAS-Elman model introduces a conditional RNN, in which the conditioner is provided by a convolutional attention-based encoder. The words-lvt5k-1sent model is also a RNN-based attention model which implements a large-vocabulary trick. Besides, RNN+MRT employs the minimum risk training strategy which directly optimizes model parameters in sentence level with respect to the evaluation metrics. SEASS (beam) extends the sequence-to-sequence framework with a selective encoding model.
%
%
The results have demonstrated that both the topic-aware module and the reinforcement learning process can improve the accuracy on text summarization. Moreover, our proposed model exhibits best scores of RG-1, RG-2 and RG-L. 

In addition, \cite{zhou2017selective} further selects 2000 pairs of summaries as an internal test set of Gigaword. We also evaluate our proposed model on this set and present the results in Table \ref{tab:res_giga2}. 
Again, our proposed model achieves the best performance in terms of all the three ROUGE scores.

\begin{table}[t]
    \centering
	\resizebox{\columnwidth}{!}{
	{
	\begin{tabular}{l|c|c|c}
		\toprule
        \hline
		 & \textbf{RG-1 (R)} & \textbf{RG-2 (R)} & \textbf{RG-L (R)}\\
		\hline
        ABS \cite{rush2015neural} & 26.55 & 7.06 & 22.05 \\
		ABS+ \cite{rush2015neural} & 28.18 & 8.49 & 23.81 \\
		RAS-Elman \cite{chopra2016abstractive} & 28.97 & 8.26 & 24.06\\
		words-lvt5k-1sent \cite{nallapati2016abstractive} & 28.61 &9.42 & 25.24\\ 
        RNN+MLE \cite{shen2016neural}  &24.92 & 8.60 & 22.25 \\
        RNN+MRT \cite{shen2016neural}  &30.41 & \textbf{10.87} & 26.79 \\
        SEASS (beam) \cite{zhou2017selective}  &29.21 & 9.56 & 25.51 \\
        ConvS2S \cite{gehring2017convolutional} &30.44 & 10.84 & 26.90 \\
        \hline
        Topic-ConvS2S & 31.08 & 10.82 & 27.61 \\
        Reinforced-ConvS2S & 30.74 & 10.68 & 27.09 \\
        Reinforced-Topic-ConvS2S & \textbf{31.15} & 10.85 & \textbf{27.68} \\
        \hline
		\bottomrule
	\end{tabular}
	}}
    \caption{Accuracy on the DUC-2004 dataset in terms of the recall-only RG-1, RG-2, and RG-L. Best performance on each score is displayed in \textbf{boldface}.}
    \label{tab:res_duc}
\end{table}

To further demonstrate the improvement of readability and diversity by the topic information, we also present some qualitative results by randomly extracting several summaries from test. We compare the reference summaries to the summaries generated by our proposed model with or without topic-aware mechanism. The examples are presented in Table \ref{tab:sum_giga}. 
We can observe that when the topic model is adopted, it can generate some accurately delivered topic words which are not in the reference summaries or the original texts. It is believed that the joint learning with a pre-trained topic model can offer more insightful information and improve the diversity and readability for the summarization.
%
%

\subsection{DUC-2004 Dataset}
Since the DUC-2004 dataset is an evaluation-only dataset, we train the models on the Gigaword corpus first and then evaluate their performance on the DUC dataset. 
As the standard practice, we report the recall-based scores of the RG-1, RG-2, and RG-L metrics in this experiment, which are given in Table \ref{tab:res_duc}.
From Table \ref{tab:res_duc} we can observe that the proposed Reinforced-Topic-ConvS2S model achieves best scores of the RG-1 and RG-L metrics, and is comparable on the RG-2 score. 
Due to the similarity of the two datasets, we do not provide qualitative summarization examples in this experiment.

\begin{table}[t]
    \centering
	\resizebox{\columnwidth}{!}{
	{\LARGE
	\begin{tabular}{l|c|c|c}
		\toprule
        \hline
		 & \textbf{RG-1 (F)} & \textbf{RG-2 (F)} & \textbf{RG-L (F)}\\
		\hline
        character-based preprocessing\\
        \hline
        RNN context \cite{hu2015lcsts} & 29.90 & 17.40 & 27.20 \\
		COPYNET \cite{gu2016incorporating} & 34.40 & 21.60 & 31.30 \\
        RNN+MLE \cite{shen2016neural} & 34.90 & 23.30 & 32.70\\
		RNN+MRT \cite{shen2016neural} & 38.20 & 25.20 & 35.40\\
        \hline
        \hline
        word-based preprocessing\\
        \hline
        RNN context \cite{hu2015lcsts} &26.80 & 16.10 & 24.10 \\       
        COPYNET \cite{gu2016incorporating}  &35.00  &22.30   &32.00   \\ 
        Topic-ConvS2S & 38.94/44.42 &21.05/32.65 & 37.03/42.09 \\
        Reinforced-ConvS2S &36.68/42.61   &18.69/29.79    &34.85/40.03  \\
        Reinforced-Topic-ConvS2S &39.93/45.12   &21.58/33.08   &37.92/42.68\\
        \hline     
		\bottomrule
	\end{tabular}
	}}
    \caption{Accuracy on the LCSTS dataset in terms of the full-length RG-1, RG-2, and RG-L. In last three rows, the word-level ROUGE scores are presented on the left and the character-level on the right.}
    \label{tab:res_lcsts}
\end{table}

\subsection{LCSTS Dataset}
We now consider the abstractive summarization task on the LCSTS dataset. Since this is a large-scale Chinese dataset, suitable data preprocessing approaches should be proposed first. Basically, there are two approaches to preprocessing the Chinese dataset: character-based and word-based. The former takes each Chinese character as the input, while the latter splits an input sentence into Chinese words. \cite{hu2015lcsts} provides a baseline result on both preprocessing approaches. 
\cite{shen2016neural} also conducts experiments on the LCSTS corpus based on character inputs. 
\cite{gu2016incorporating} proposes a neural model, the COPYNET, with both character-based and word-based preprocessing by incorporating the copying mechanism into the sequence-to-sequence framework.
In this work, we adopt the word-based approach as we believe that in the case of Chinese, words are more relevant to latent knowledge of documents than characters are.

Since the standard ROUGE package\footnote{\url{http://www.berouge.com/Pages/default.aspx}} is usually used to evaluate the English summaries, directly employing the package to evaluate Chinese summaries would yield underrated results. In order to evaluate the summarization on the LCSTS dataset, we follow the suggestion of \cite{hu2015lcsts} by mapping Chinese words/characters to numerical IDs, on which we then perform the ROUGE evaluation. Since not all previous work explicitly mentioned whether word-based or character-based ROUGE metrics were reported, we evaluate our proposed model with both metrics in order to obtain a comprehensive comparison. The results of both scores are presented in Table \ref{tab:res_lcsts}, which are displayed as word-based score/character-based score.

From the results shown in Table \ref{tab:res_lcsts}, we see that one can always achieve higher ROUGE scores in the character level than that based on Chinese words by our proposed model. We can also observe that the character-based results of our Reinforced-Topic-ConvS2S model outperforms every other method. Regarding to word-based ROUGE scores, our model obtains the best performance in terms of RG-1 and RG-L metrics. 
However, our best model does not achieve a good RG-2 score as its RG-1 and RG-L scores.
We suspect that it may be partly caused by the biased probability generation mechanism that influences word order, which requires further studies.


In addition to ROUGE scores, we also present some randomly picked examples of generated summaries in Table \ref{tab:sum_lcsts}. The original examples (in Chinese) are shown and all the texts are carefully translated to English for the convenience of reading. The examples demonstrate that the topic-aware mechanism can also improve the diversity in Chinese summarization tasks.

\section{Conclusion and Future Work}
In this work, we propose a topic-aware ConvS2S model with reinforcement learning for abstractive text summarization. It is demonstrated that the new topic-aware attention mechanism introduces some high-level contextual information for summarization. The performance of the proposed model advances state-of-the-art methods on various benchmark datasets. In addition, our model can produce summaries with better informativeness, coherence, and diversity.

Note that the experiments in this work are mainly based on the sentence summarization. In the future, we aim to evaluate our model on the datasets where the source texts can be long paragraphs or multi-documents. 
Moreover, we also note that how to evaluate the performance on Chinese summaries remains an open problem. It is also of great interest to study on this subject in the future. 

\section*{Acknowledgements}
Qiang Du is supported in part by the US NSF TRIPODs project through CCF-170483.

\begin{CJK*}{UTF8}{gbsn}
\begin{table*}[ht]
	\centering
	\resizebox{2\columnwidth}{!}{
	{\tiny
	\begin{tabular}{m{11cm}}
		\toprule
        \hline
		 \textbf{Examples of summaries} \\
         \hline         
         $\bm{D}$: 根据 \#\#\#\# 年 \# 月 \# 日 国家 发改委 等 部门 联合 发布 的 《 关于 进一步 做好 新能源 汽车 推广应用 工作 的 通知 》， \#\#\#\# 年 的 补贴 金额 相比 \#\#\#\# 年 将 降低 \#\#\% 。 （ 分享 自 @ 电动 邦 ）\\
         $\bm{D}$: According to the notice {\em On the further promotion and application of new energy vehicles}, jointly released by the National Development and Reform Commission and other departments on \#\#/\#\#/\#\#\#\# (date), the compensation of \#\#\#\# (year) will be reduced by \#\#\% compared to \#\#\#\# (year). (reposted from @electric\_nation) \\
         $\bm{R}$: 补贴 金额 再 缩水 \#\#\#\# 年 新能源 车 政策 解读 \\
         $\bm{R}$: The compensation has been reduced again: \#\#\#\# (year) policy analysis of new energy automobiles\\
         $\bm{OR}$: \#\#\#\# 年 新能源 汽车 推广应用 工作 的 通知\\
         $\bm{OR}$: \#\#\#\# (year) notice on the promotion and application of new energy vehicles\\
         $\bm{OT}$: {\color{blue}国家 发改委} {\color{red}发文} {\color{blue}进一步 做好} 新能源 {\color{blue}汽车} 推广应用 {\color{blue}工作}\\
         $\bm{OT}$: The {\color{blue}National Development and Reform Commission} {\color{red}issued a policy} on {\color{blue}further} promotion and application of new energy {\color{blue}vehicles}\\
         \hline
         
         $\bm{D}$: 成都市 软件 和 信息技术 服务业 近年来 一直 保持 快速 增长势头 ， 稳居 中西部 城市 之 首 ， 已 成为 我国 西部 “ 硅谷 ” 。 《 \#\#\#\# 年度 成都市 软件 和 信息技术 服务 产业 发展 报告 》 日前 发布 … … 详情请 见 : @ 成都日报 @ 成都 发布\\
         $\bm{D}$: In recent years, the service industry of software and information technology in Chengdu has been growing rapidly, ranking first among the cities in Midwest China. Chengdu has become China's western ``Silicon Valley''. The {\em \#\#\#\# (year) Annual Chengdu Software and Information Technology Service Industry Development Report} has been released recently ... ... see details: @ Chengdu\_Daily @ Chengdu\_release \\
         $\bm{R}$: 成都 倾力 打造 西部 “ 硅谷 ” \\
         $\bm{R}$: Chengdu makes every effort to build the western ``Silicon Valley'' \\
         $\bm{OR}$: 成都 {\color{blue}软件} 和 信息技术 {\color{blue}服务业 发展 报告 发布}\\
         $\bm{OR}$: The {\color{blue}report} of Chengdu {\color{blue}software} and information technology {\color{blue}service industry development} has been {\color{blue}released}\\
         $\bm{OT}$: 成都 {\color{blue}软件} 和 信息技术 {\color{blue}服务业} {\color{red}跃居} 西部 “ 硅谷 ”\\
         $\bm{OT}$: The {\color{blue}service industry} of {\color{blue}software} and information technology in Chengdu {\color{red}rockets to} make it the western ``Silicon Valley'' \\
         \hline
         
         $\bm{D}$: 新疆 独特 的 区位 优势 ， 使 其 成为 “ 一带 一路 ” 战略 重要一环 。 记者 从 新疆 发改委 获悉 ， 库尔勒 至 格尔木 铁路 先期 开工 段 已 进入 招投标 阶段 ， 计划 \#\#\#\# 年 \#\# 月 中旬 正式 开工 建设 。 \#\#\#\# 年 计划 完成 投资 \#\# 亿元 。\\
         $\bm{D}$: Xinjiang's unique geographical advantages make it an important part of {\em The Belt and Road} strategy. The reporter learned from the Xinjiang Development and Reform Commission that the initial railway construction project from Korla to Golmud had been on tendering procedure. The project was scheduled to officially launch in mid \#\# (month) of \#\#\#\# (year) and attract the investment of \#\# billion yuan by \#\#\#\# (year). \\
         $\bm{R}$: “ 一带 一路 ” 战略 惠及 新疆 $<$unk$>$, 铁路 年底 开建 \\
         $\bm{R}$: {\em The Belt and Road} strategy benefits Xinjiang $<$unk$>$ and the railway construction starts by the end of \#\#\#\# (year) \\
         $\bm{OR}$: 新疆 $<$unk$>$ 至 格尔木 铁路 计划 \#\#\#\# 年 开建 \\
         $\bm{OR}$: The railway from $<$unk$>$ to Golmud is scheduled to start construction in \#\#\#\# (year)\\
         $\bm{OT}$: 库尔勒 至 格尔木 铁路 {\color{red}拟} \#\# 月 {\color{blue}开工 建设}\\
         $\bm{OT}$: The railway construction {\color{blue}project} from Korla to Golmud {\color{red}is planned} to {\color{blue}launch} in \#\# (month)\\
         \hline
         
         
         $\bm{D}$: 昨日 ， 商报 记者 从 代表 国内 婚尚 产业 “ 风向标 ” 的 上海 国际 婚纱 摄影 器材 展览会 上 了解 到 ， 部分 商家 开始 将 婚庆 布置 、 婚礼 流程 、 形式 交给 新人 决定 以 迎合 \#\# 后 新人 的 需求 。 此次 展览会 的 规模 超过 \# 万平方米 ， 吸引 参展 企业 超过 \#\#\# 家 。\\
         $\bm{D}$: The day before, the reporters of {\em Commercial News} learned from the Shanghai International Wedding Photographic Equipment Exhibition, which has been leading and defining the domestic wedding industry, that some companies began to cater for the requirements of \#\#s-generation newly married couples by self-decided wedding decoration, wedding process and forms. The venue of the exhibition is more than \# tens of thousands square meters, attracting more than \#\#\# exhibitors.\\
         $\bm{R}$: 婚庆 “ 私人 定制 ” 受 \#\# 后 新人 追捧 \\
         $\bm{R}$: The personalized wedding is admired by \#\#s-generation newly married couples \\
         $\bm{OR}$: {\color{blue}上海} 国际 婚纱 {\color{blue}摄影} 器材 展览会 举行\\
         $\bm{OR}$: {\color{blue}Shanghai} International Wedding {\color{blue}Photographic} Equipment Exhibition was held\\
         $\bm{OT}$: {\color{blue}上海} 国际 婚纱 {\color{blue}摄影} 器材 展览会 {\color{red}昨} 举行\\
         $\bm{OT}$: {\color{blue}Shanghai} International Wedding {\color{blue}Photographic} Equipment Exhibition was held {\color{red} yesterday}\\
         \hline
         
		\bottomrule
	\end{tabular}
	}}
    \caption{Examples of generated summaries on the LCSTS dataset. \textbf{D}: source document, \textbf{R}: reference summary, \textbf{OR}: output of the \textbf{R}einforced-ConvS2S model, \textbf{OT}: output of the Reinforced-\textbf{T}opic-ConvS2S model. The words marked in {\color{blue}blue} are topic words not in the reference summaries. The words marked in {\color{red}red} are topic words neither in the reference summaries nor in the source documents. All the texts are carefully \textit{translated from Chinese}.}
    \label{tab:sum_lcsts}
	\vspace{8pt}  
\end{table*}
\clearpage\end{CJK*}

\newpage
\bibliographystyle{named}
\bibliography{textsum}

\end{document}